\documentclass{article}

\usepackage{arxiv}

\usepackage[utf8]{inputenc} 
\usepackage[T1]{fontenc}    
\usepackage{hyperref}       
\usepackage{url}            
\usepackage{booktabs}       
\usepackage{amsfonts}       
\usepackage{nicefrac}       
\usepackage{microtype}      
\usepackage{lipsum}         
\usepackage{graphicx}
\usepackage{natbib}
\usepackage{doi}

\usepackage{amsmath,amssymb,amsthm}
\usepackage{algorithm}
\usepackage{algorithmic}
\usepackage{booktabs}

\newtheorem{definition}{Definition}

\usepackage{tikz}
\usetikzlibrary{arrows.meta, positioning}

\title{How Hard is it to Confuse a World Model?}

\date{}

\newif\ifuniqueAffiliation
\uniqueAffiliationtrue

\ifuniqueAffiliation 
\author{ 
    \href{https://orcid.org/0009-0001-9818-0171}{\includegraphics[scale=0.06]{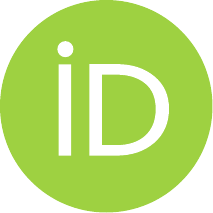}\hspace{1mm}Waris Radji} \quad
    \href{https://orcid.org/0000-0001-7935-7026}{\includegraphics[scale=0.06]{orcid.pdf}\hspace{1mm}Odalric-Ambrym Maillard} \\
    Univ. Lille, Inria, CNRS, Centrale Lille, UMR 9189-CRIStAL, F-59000 Lille, France \\
    \texttt{\{waris.radji,odalric.maillard\}@inria.fr}
}
\else
\usepackage{authblk}

\newbox{\orcid}\sbox{\orcid}{\includegraphics[scale=0.06]{orcid.pdf}} 
\author[1]{%
	\href{https://orcid.org/0000-0000-0000-0000}{\usebox{\orcid}\hspace{1mm}David S.~Hippocampus\thanks{\texttt{hippo@cs.cranberry-lemon.edu}}}%
}
\author[1,2]{%
	\href{https://orcid.org/0000-0000-0000-0000}{\usebox{\orcid}\hspace{1mm}Elias D.~Striatum\thanks{\texttt{stariate@ee.mount-sheikh.edu}}}%
}
\affil[1]{Department of Computer Science, Cranberry-Lemon University, Pittsburgh, PA 15213}
\affil[2]{Department of Electrical Engineering, Mount-Sheikh University, Santa Narimana, Levand}
\fi

\renewcommand{\headeright}{}
\renewcommand{\undertitle}{}
\renewcommand{\shorttitle}{}

\begin{document}
\maketitle

\begin{abstract}
In reinforcement learning (RL) theory, the concept of most confusing instances is central to establishing regret lower bounds, that is, the minimal exploration needed to solve a problem. Given a reference model and its optimal policy, a most confusing instance is the statistically closest alternative model that makes a suboptimal policy optimal. While this concept is well-studied in multi-armed bandits and ergodic tabular Markov decision processes, constructing such instances remains an open question in the general case. In this paper, we formalize this problem for neural network world models as a constrained optimization: finding a modified model that is statistically close to the reference one, while producing divergent performance between optimal and suboptimal policies. We propose an adversarial training procedure to solve this problem and conduct an empirical study across world models of varying quality. Our results suggest that the degree of achievable confusion correlates with uncertainty in the approximate model, which may inform theoretically-grounded exploration strategies for deep model-based RL.
\end{abstract}

\section{Introduction}

Model-based reinforcement learning (RL) achieves superior sample efficiency by learning dynamics models that enable planning and simulation \citep{chua2018deep,janner2019trust,hansen2023td,hafner2025training}. But a fundamental challenge remains: \textit{how should agents explore when their world models are uncertain?} Existing deep RL methods rely on heuristic uncertainty quantification, including, for example, ensemble disagreement \citep{pathak2019self,shyam2019model}, prediction error bonuses \citep{pathak2017curiosity}, and variational inference in Bayesian neural networks \citep{houthooft2016vime}. These strategies that have good empirical results lack direct connections to the requirements of optimal exploration.
Multi-armed bandits (MABs) offer a precise characterization of exploration difficulty through regret lower bounds: optimal algorithms must accumulate sufficient statistical evidence to distinguish the true environment from \emph{confusing instances}—plausible alternative models that are statistically similar to the true model yet require different optimal actions \citep{10.1016/0196-8858(85)90002-8, Burnetas1996122,06a276ef-3e35-33d0-912a-626bc0841b5e}. 
Exploration strategies that explicitly construct and resolve such confusing instances achieve \textit{asymptotically} optimal regret in various settings \citep{honda2011asymptotically,pesquerel2022imed}. However, finding these instances in practice, especially for continuous or high-dimensional domains with learned models, remains unexplored and NP-Hard \citep{radjiconfusing,boone2025regret}.
\paragraph{Research problem.} The central computational challenge is identifying the \emph{most} confusing instance: the alternative model that is statistically closest to the learned model while still inducing a different optimal policy. For general Markov decision processes, this search is NP-hard, making optimal exploration seemingly intractable in high-dimensional state spaces.
In this paper, we investigate this problem empirically for probabilistic world models based on variational autoencoder (VAE) architectures \citep{ha2018world}. We formalize the search for the most confusing instance as a constrained optimization problem, via Lagrangian relaxation, introducing the \emph{suboptimality cost}, that is, the minimal log-likelihood ratio required to reverse policy rankings between the learned model and a confusing alternative. Our key empirical finding is a relationship between model quality and confusion resistance: better-trained models have higher suboptimality costs. This intuitive relationship suggests that suboptimality cost is related to model uncertainty, offering a promising approach to quantifying exploration difficulty that paves the way for new exploration strategies in deep model-based RL.


\paragraph{Background and setting.}
We consider episodic reinforcement learning in continuous state and action spaces $\mathcal{S}$ and $\mathcal{A}$ with reward function $R: \mathcal{S} \times \mathcal{A} \to \mathbb{R}$ \citep{sutton1998reinforcement}. A learned world model $M_\theta$ with parameters $\theta$ approximates environment dynamics through probabilistic transitions $P_\theta(s_{t+1} | s_t, a_t)$, trained on observed environment data. We assume these transitions are differentiable with respect to $\theta$, enabling gradient-based optimization.
Given a starting state $s_0$, let $V_\pi(s_0; M)$ denote the expected return of policy $\pi$ under model $M$, as
\begin{equation}
V_\pi(s_0; M) = \mathbb{E}_{\tau \sim M,\pi}\left[\sum_{t=0}^{T-1} R(s_t, a_t)\right],
\end{equation}
where $\tau = (s_0, a_0, s_1, \ldots, s_{T-1}, a_{T-1})$ is a trajectory of fixed horizon $T$.
Consider $\pi^*$ optimal under the reference model $M_\theta$, and $\pi_{\text{sub}}$ suboptimal under $M_\theta$, meaning $V_{\pi^*}(s_0; M_\theta) > V_{\pi_{\text{sub}}}(s_0; M_\theta)$.

\section{Most Confusing Instance Search}

\begin{definition}[Confusing Instance]
\label{def:confusing}
Given reference model $M_\theta$ with optimal policy $\pi^*$ and suboptimal policy $\pi_{\text{sub}}$ at state $s_0$, a model $\widetilde{M} = M_{\tilde{\theta}}$ is a \emph{confusing instance} if it is statistically close to $M_\theta$ yet inverts the policy ranking:
\begin{equation}
V_{\pi^*}(s_0; \widetilde{M}) < V_{\pi_{\text{sub}}}(s_0; \widetilde{M}), \
\text{and KL}_T^{\pi^*}(M_\theta \| M_{\tilde{\theta}}) = \underset{{\tau \sim M_\theta, \pi^*}}{\mathbb{E}}\left[\sum_{t=0}^{T-1} \log \frac{P_\theta(s_{t+1} | s_t, a_t)}{P_{\tilde{\theta}}(s_{t+1} | s_t, a_t)}\right], \text{ is small}.
\end{equation}
\end{definition}
The last quantity represents the expected log-likelihood ratio of trajectories generated by $\pi^*$ under the two models. A small (minimal) divergence implies we cannot reliably distinguish between $M_\theta$ and $\widetilde{M}$ by executing only the optimal policy; yet the models prescribe different optimal behaviors.

\begin{definition}[Suboptimality Cost]
\label{def:subopt_cost}
 Among all confusing instances, we seek the one that is statistically closest to the reference model. The \emph{suboptimality cost} is the infimum of KL divergences over all confusing instances:
\begin{equation}
\label{eq:subopt_cost}
\mathbf{K}(s_0, \pi^*, \pi_{\text{sub}}) = \inf_{\tilde{\theta}} \left\{ \text{KL}_T^{\pi^*}(M_\theta \| M_{\tilde{\theta}}) : V_{\pi^*}(s_0; \widetilde{M}) < V_{\pi_{\text{sub}}}(s_0; \widetilde{M}) \right\}.
\end{equation}
\end{definition}
This cost quantifies the minimal statistical evidence needed to distinguish which policy is truly superior. RL theory indicates that algorithms must accumulate at least $\mathbf{K}$ nats to avoid linear regret. Thus, $\mathbf{K}$ measures model uncertainty for exploration: low values indicate the model is easily confused and requires more data, while high values indicate alternative explanations are statistically distant, making the model reliable for decision-making.

\paragraph{Optimization.} Finding \eqref{eq:subopt_cost} is a challenging constrained optimization: the objective involves expectations over stochastic trajectories, and the constraint depends on value functions that themselves require trajectory rollouts. While this problem is generally NP-hard, the differentiable structure of neural world models enables tractable local optimization via gradient descent. We address these challenges through Lagrangian relaxation and sampling-based gradient estimation: 
\begin{equation}
\mathcal{L}(\tilde{\theta}, \lambda) = \text{KL}_T^{\pi^*}(M_\theta \| M_{\tilde{\theta}}) + \lambda \max\left(0, V_{\pi^*}(M_{\tilde{\theta}}) - V_{\pi_{\text{sub}}}(M_{\tilde{\theta}})\right)
\end{equation}
where $\lambda \geq 0$ is the Lagrange multiplier. The algorithm alternates between:
\begin{align}
\text{(Primal)} \quad \tilde{\theta}^{(k+1)} &\leftarrow \tilde{\theta}^{(k)} - \alpha_\theta \nabla_{\tilde{\theta}} \mathcal{L}(\tilde{\theta}^{(k)}, \lambda^{(k)}) \label{eq:primal}\\
\text{(Dual)} \quad \lambda^{(k+1)} &\leftarrow \max\left(0, \lambda^{(k)} + \alpha_\lambda \cdot c(\tilde{\theta}^{(k)})\right) \label{eq:dual}
\end{align}
where $c(\tilde{\theta}) = V_{\pi^*}(M_{\tilde{\theta}}) - V_{\pi_{\text{sub}}}(M_{\tilde{\theta}})$.
When the constraint is violated ($c > 0$), the multiplier $\lambda$ increases, penalizing the objective more heavily. When satisfied ($c \leq 0$), we record the current KL as a candidate for $\mathbf{K}$. The infimum is the smallest KL achieved among all feasible iterates. An analogy can be made with policy gradient algorithms with KL regularization \citep{schulman2015trust,schulman2017proximal}: instead of optimizing the policy, we optimize a world model.

\section{Numerical Experiments}

While our framework applies to any differentiable probabilistic model, we instantiate it for variational autoencoder world models in our experiments due to their straightforward implementation and interpretable latent representations. A VAE world model decomposes as $M_\theta = (\phi, \psi, \xi)$, where $\phi$ is an encoder mapping states to latent codes $z \sim q_\phi(z|s)$, $\psi$ models the latent dynamics $p_\psi(z'|z,a)$, and $\xi$ is a decoder reconstructing states $s \sim p_\xi(s|z)$. We focus on fully observable environments in our experiments, so our model uses a feedforward architecture without recurrent components.

\begin{algorithm}[h]
\caption{Compute Suboptimality Cost in VAE}
\label{alg:subopt_cost}
\begin{algorithmic}[1]
\REQUIRE Reference model $M_\theta =  (\phi, \psi, \xi)$, policies $\pi^*, \pi_{\text{sub}}$, horizon $T$
\ENSURE Suboptimality cost $\mathbf{K}$
\STATE Initialize: $\tilde{\psi} \leftarrow \psi$, $\lambda \leftarrow 1.0$, $\mathbf{K}_{\text{best}} \leftarrow +\infty$
\FOR{$k = 1, \ldots, K$}
    \STATE Sample $N$ initial states: $s_0^{(n)} \sim p_0$ for $n=1,\ldots,N$
    \STATE Encode: $z_0^{(n)} = \phi(s_0^{(n)})$ (deterministic for frozen encoder)
    \FOR{$n = 1, \ldots, N$}
        \STATE Rollout $\pi^*$ in $M_{\tilde{\theta}} = (\phi, \tilde{\psi}, \xi)$: generate trajectory $\tau^{*(n)} = \{(z_t, a_t, r_t)\}_{t=0}^{T-1}$
        \STATE Rollout $\pi_{\text{sub}}$ in $M_{\tilde{\theta}}$: generate trajectory $\tau_{\text{sub}}^{(n)}$
    \ENDFOR
    \STATE Compute average KL: $\overline{\text{KL}} = \frac{1}{NT}\sum_{n=1}^N \sum_{t=0}^{T-1} \text{KL}(p_\psi(\cdot|z_t^{(n)}, a_t^{(n)}) \| p_{\tilde{\psi}}(\cdot|z_t^{(n)}, a_t^{(n)}))$
    \STATE Compute average returns: $\overline{V^*} = \frac{1}{N}\sum_{n} \sum_{t} r_t^{*(n)}$, $\overline{V_{\text{sub}}} = \frac{1}{N}\sum_{n} \sum_{t} r_t^{\text{sub},(n)}$
    \STATE Constraint violation: $c = \overline{V^*} - \overline{V_{\text{sub}}}$
    \IF{$c \leq 0$}
        \STATE $\mathbf{K}_{\text{best}} \leftarrow \min(\mathbf{K}_{\text{best}}, \overline{\text{KL}})$ \COMMENT{Record feasible KL}
    \ENDIF
    \STATE Update dynamics: $\tilde{\psi} \leftarrow \tilde{\psi} - \alpha_\psi \nabla_{\tilde{\psi}}[\overline{\text{KL}} + \lambda \cdot \max(0, c)]$
    \STATE Update multiplier: $\lambda \leftarrow \max(0, \lambda + \eta \cdot c)$
\ENDFOR
\RETURN $\mathbf{K}_{\text{best}}$
\end{algorithmic}
\end{algorithm}

\subsection{World Model Architecture and Training}

Our VAE-based world model consists of three components:
\begin{itemize}
    \item \textbf{Encoder} $\phi$: Two-layer MLP with hidden dimension 64 and tanh activations, mapping observations $s \in \mathbb{R}^4$ to latent codes $z \in \mathbb{R}^{d_z}$ where $d_z \in \{32, 64\}$.
    \item \textbf{Dynamics} $\psi$: Two-layer MLP with hidden dimension 64, taking concatenated latent-action pairs $(z, a) \in \mathbb{R}^{d_z+2}$ and outputting Gaussian parameters $(\mu_\psi, \sigma_\psi^2)$ for the next latent state. We parameterize variance as $\sigma^2 = (\text{softplus}(\text{logvar}_{\text{raw}}) + 10^{-8})^2$ for numerical stability.
    \item \textbf{Decoder} $\xi$: Two-layer MLP with hidden dimension 64 and tanh activations, reconstructing state differences $\Delta s \in \mathbb{R}^4$ from latent codes.
\end{itemize}

We train on 50,000 transitions collected via random exploration (500 trajectories of 200 steps each, with action repeat sampled uniformly). The loss combines reconstruction and KL regularization:
\begin{equation}
\mathcal{L}_{\text{train}} = \mathbb{E}_{(s,a,\Delta s) \sim \mathcal{D}}\left[\|\Delta \hat{s} - \Delta s\|^2\right] + 0.01 \cdot \text{KL}(q_\phi(\cdot|s) \| \mathcal{N}(0, I))
\end{equation}
Training proceeds for 500 epochs with the Adam optimizer (learning rate $10^{-3}$) and batch size 256. We save checkpoints at epochs 100, 200, 300, 400, and 500, providing a spectrum of model qualities from under-trained (high uncertainty) to well-trained (low uncertainty).

\subsection{Confusing Instance Search Implementation}

To isolate dynamical uncertainty from representational artifacts, we \textit{freeze} the encoder $\phi$ and decoder $\xi$, perturbing only the latent dynamics $\psi$. This ensures the confusing instance operates in the same representational space, making comparisons meaningful, and avoiding trivial confusion of the model by distorting the representation itself.
For Gaussian latent dynamics $p_\psi(z'|z,a) = \mathcal{N}(\mu_\psi(z,a), \text{diag}(\sigma_\psi^2(z,a)))$, the pointwise KL divergence has closed form:
\begin{equation}
\text{KL}(p_\psi \| p_{\tilde{\psi}}) = \frac{1}{2}\sum_{i=1}^{d_z}\left[\log \frac{\tilde{\sigma}_{\psi,i}^2}{\sigma_{\psi,i}^2} + \frac{\sigma_{\psi,i}^2 + (\mu_{\psi,i} - \tilde{\mu}_{\psi,i})^2}{\tilde{\sigma}_{\psi,i}^2} - 1\right]
\end{equation}
We approximate trajectory-level KL by averaging pointwise KL along rollouts:
\begin{equation}
\text{KL}_T^{\pi^*}(M_\theta \| M_{\tilde{\theta}}) \approx \mathbb{E}_{\tau \sim M_{\tilde{\theta}}, \pi^*}\left[\frac{1}{T}\sum_{t=0}^{T-1} \text{KL}(p_\psi(\cdot | z_t, a_t) \| p_{\tilde{\psi}}(\cdot | z_t, a_t))\right]
\end{equation}
We optimize confusing instances for 2500 iterations with $N=128$ trajectory samples per iteration. The learning rate for dynamics parameters follows a linear schedule from $5 \times 10^{-4}$ to $1 \times 10^{-4}$, with gradient clipping at max norm 1.0. The Lagrange multiplier $\lambda$ is initialized to 1.0 and updated with step size $\eta = 0.1$. We use the Adam optimizer and save checkpoints every 100 iterations. Algorithm~\ref{alg:subopt_cost} presents the complete procedure for computing suboptimality cost.

\subsection{Task and Policy Construction}

\begin{figure}[h]
    \centering
    \begin{tikzpicture}[scale=0.8]
        \begin{scope}[xshift=0cm]
            \node[above] at (2.5, 5.2) {\textbf{Original Maze}};
            \node[above, font=\small] at (2.5, 4.8) {};
            
            \draw[step=1cm, gray, very thin] (0,0) grid (5,5);
            
            \fill[black] (0,0) rectangle (5,1);
            \fill[black] (0,1) rectangle (1,2);
            \fill[black] (4,1) rectangle (5,2);
            \fill[black] (0,2) rectangle (1,3);
            \fill[black] (1,2) rectangle (2,3);
            \fill[black] (2,2) rectangle (3,3);
            \fill[black] (4,2) rectangle (5,3);
            \fill[black] (0,3) rectangle (1,4);
            \fill[black] (4,3) rectangle (5,4);
            \fill[black] (0,4) rectangle (5,5);
            
            \node[circle, fill=green!30, inner sep=3pt] at (1.5, 1.5) {S};
            
            \node[circle, fill=red!30, inner sep=3pt] at (1.5, 3.5) {G};
            
            \draw[->, thick, blue!70, line width=1.5pt] 
                (1.5, 1.5) -- (2.5, 1.5) -- (3.5, 1.5) -- (3.5, 2.5) 
                -- (3.5, 3.5) -- (2.5, 3.5) -- (1.5, 3.5);
        \end{scope}
        
        \begin{scope}[xshift=7cm]
            \node[above] at (2.5, 5.2) {\textbf{Modified Maze}};
            \node[above, font=\small] at (2.5, 4.8) {};
            
            \draw[step=1cm, gray, very thin] (0,0) grid (5,5);
            
            \fill[black] (0,0) rectangle (5,1);
            \fill[black] (0,1) rectangle (1,2);
            \fill[black] (4,1) rectangle (5,2);
            \fill[black] (0,2) rectangle (1,3);
            \fill[black] (2,2) rectangle (3,3);
            \fill[black] (3,2) rectangle (4,3);
            \fill[black] (4,2) rectangle (5,3);
            \fill[black] (0,3) rectangle (1,4);
            \fill[black] (4,3) rectangle (5,4);
            \fill[black] (0,4) rectangle (5,5);
            
            \node[circle, fill=green!30, inner sep=3pt] at (1.5, 1.5) {S};
            
            \node[circle, fill=red!30, inner sep=3pt] at (1.5, 3.5) {G};
            
            \draw[->, thick, blue!70, line width=1.5pt] 
                (1.5, 1.5) -- (1.5, 2.5) -- (1.5, 3.5);
        \end{scope}
        
        \node at (6, -1) {\small 
            \begin{tikzpicture}[baseline=-0.5ex]
                \fill[black] (0,0) rectangle (0.3,0.3);
            \end{tikzpicture} Wall \quad
            \begin{tikzpicture}[baseline=-0.5ex]
                \node[circle, fill=green!30, inner sep=2pt] at (0.15,0.15) {\tiny S};
            \end{tikzpicture} Start \quad
            \begin{tikzpicture}[baseline=-0.5ex]
                \node[circle, fill=red!30, inner sep=2pt] at (0.15,0.15) {\tiny G};
            \end{tikzpicture} Goal
        };
    \end{tikzpicture}
    \caption{\textbf{Maze configurations for policy construction.} \textit{Left:} Original U-maze where $\pi^*$ follows the shortest path around the obstacle. \textit{Right:} Modified maze where the wall configuration is altered, allowing $\pi_{\text{sub}}$ to take a direct vertical path. When executed in the original maze, $\pi_{\text{sub}}$ cannot pass through the wall and must take a longer route.}
    \label{fig:maze_layouts}
\end{figure}
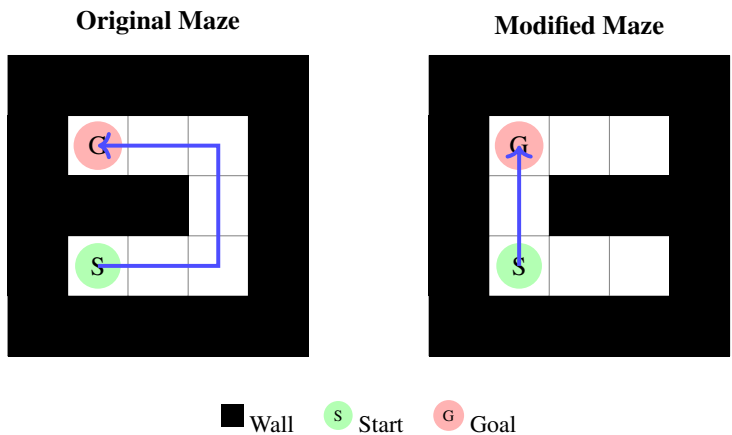

We evaluate on a U-shaped maze environment with a $5 \times 5$ grid structure from \cite{pointax2025}. The agent has a continuous state representation (2D position and velocity) and must navigate to a fixed goal location. We use a dense reward signal equal to the negative Manhattan distance between the agent's position and the goal. 
We construct two policy-maze pairs to define optimal and suboptimal behaviors. The optimal policy $\pi^*$ is computed via breadth-first search (BFS) on the original discretized U-Maze, following the shortest path. The suboptimal policy $\pi_{\text{sub}}$ is obtained by first moving one wall in the maze to create an alternative layout, then computing BFS on this modified maze. When executed in the original U-Maze, this policy follows a longer route and is thus suboptimal. Both policies have been augmented with Gaussian noise of $\mathcal{N}(0, 1)$ to introduce stochasticity during rollouts. Figure~\ref{fig:maze_layouts} shows the maze configurations used to construct the two policies.

\subsection{Experimental Results}

\begin{figure}[!t]
    \centering
\includegraphics[width=\linewidth]{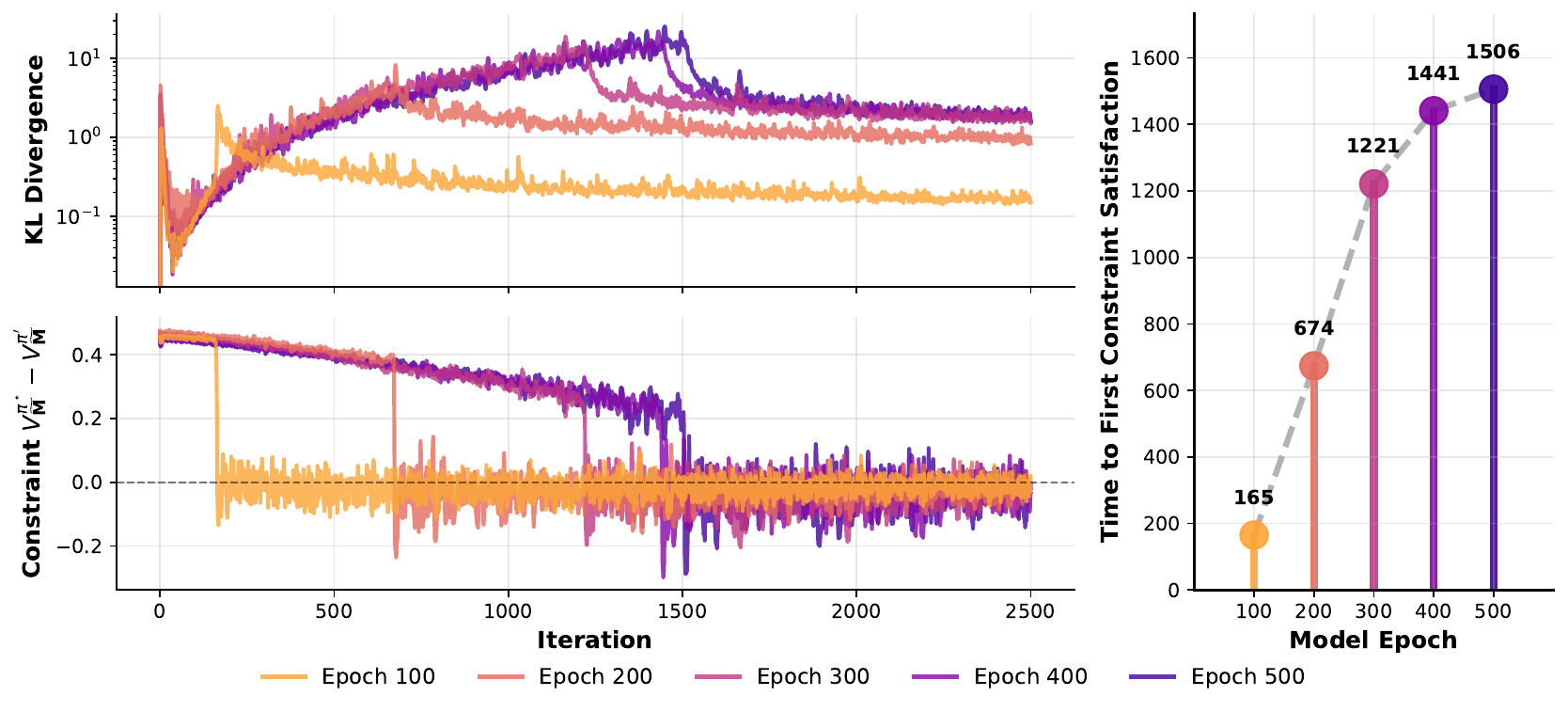}
    \caption{\textbf{Confusing instance search optimization dynamics.} \textit{Left:} KL divergence (logarithmic scale) and constraint evolution over training iterations. \textit{Right:} Time to first constraint satisfaction.}
    \label{fig:training_dynamics}
\end{figure}

\paragraph{How does model quality affect the ease of finding confusing instances?} Figure~\ref{fig:training_dynamics} reveals a clear trend: under-trained models (epochs 100-200) are easily confused, with the constraint satisfied almost immediately and remaining deeply negative throughout training. In contrast, well-trained models (epochs 300-500) exhibit strong resistance, requiring the optimizer to explore significantly higher KL values before finding confusing instances. The rightmost panel quantifies this relationship precisely: the time to first constraint satisfaction grows from 165 iterations (epoch 100) to 1506 iterations (epoch 500). Notably, this growth appears logarithmic rather than linear: the increments diminish as models improve. This sublinear scaling aligns with fundamental exploration theory: efficient algorithms should accumulate evidence at diminishing rates as they learn, achieving sublinear regret. The pattern seems to demonstrate that the suboptimality cost $\mathbf{K}$ increases monotonically with model quality, as better-trained models have learned tighter dynamics distributions that fundamentally constrain the space of statistically plausible yet behaviorally-distinct alternatives.

\begin{figure}[!b]
    \centering
    \includegraphics[width=1.\linewidth]{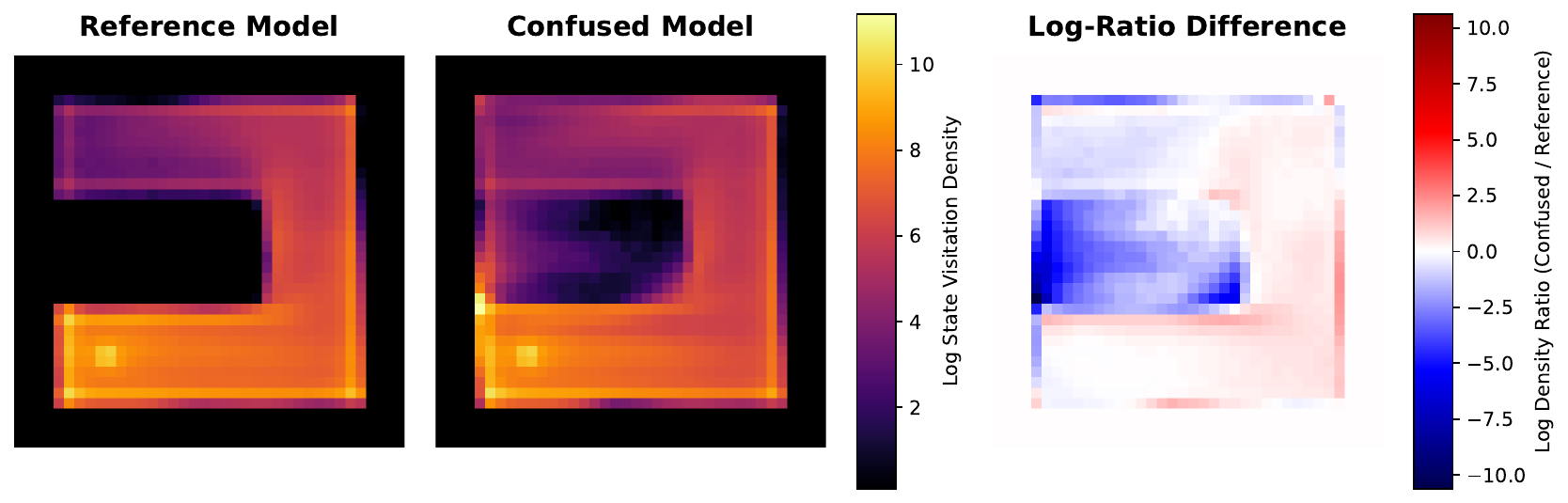}
    \caption{\textbf{Statistical closeness of confusing instances.} \textit{Left/Center:} State visitation densities (log scale) under random policy. \textit{Right:} Log-density ratio shows localized changes: blue indicates higher probability in the confusing instance, red indicates lower probability.}
    \label{fig:density_heatmap}
\end{figure}

\paragraph{How close are confusing instances to their reference models?}
Figure~\ref{fig:density_heatmap} visualizes the statistical proximity between reference and confusing models through state visitation densities obtained under a fully random policy run for 200 timesteps. We examine the confusing instance found at epoch 500, the hardest to confuse. At first glance, the left and center panels show similar overall patterns, but the log-density ratio (right panel) reveals the mechanism of confusion through \emph{localized} redistributions. Blue regions indicate where the confused model assigns \emph{higher} visitation probability. These correspond to wall regions where the confused model effectively enables passage, creating shortcuts that make the suboptimal policy viable. Red regions show where the confused model assigns a lower probability. Crucially, most of the state space remains white, indicating preserved statistical similarity. The confusing instance achieves behavioral change through minimal, targeted perturbations: it exploits the reference model's uncertainty about wall dynamics precisely where shortcuts matter, while maintaining near-identical predictions elsewhere.

\begin{figure}[h]
    \centering
    \includegraphics[width=1.\linewidth]{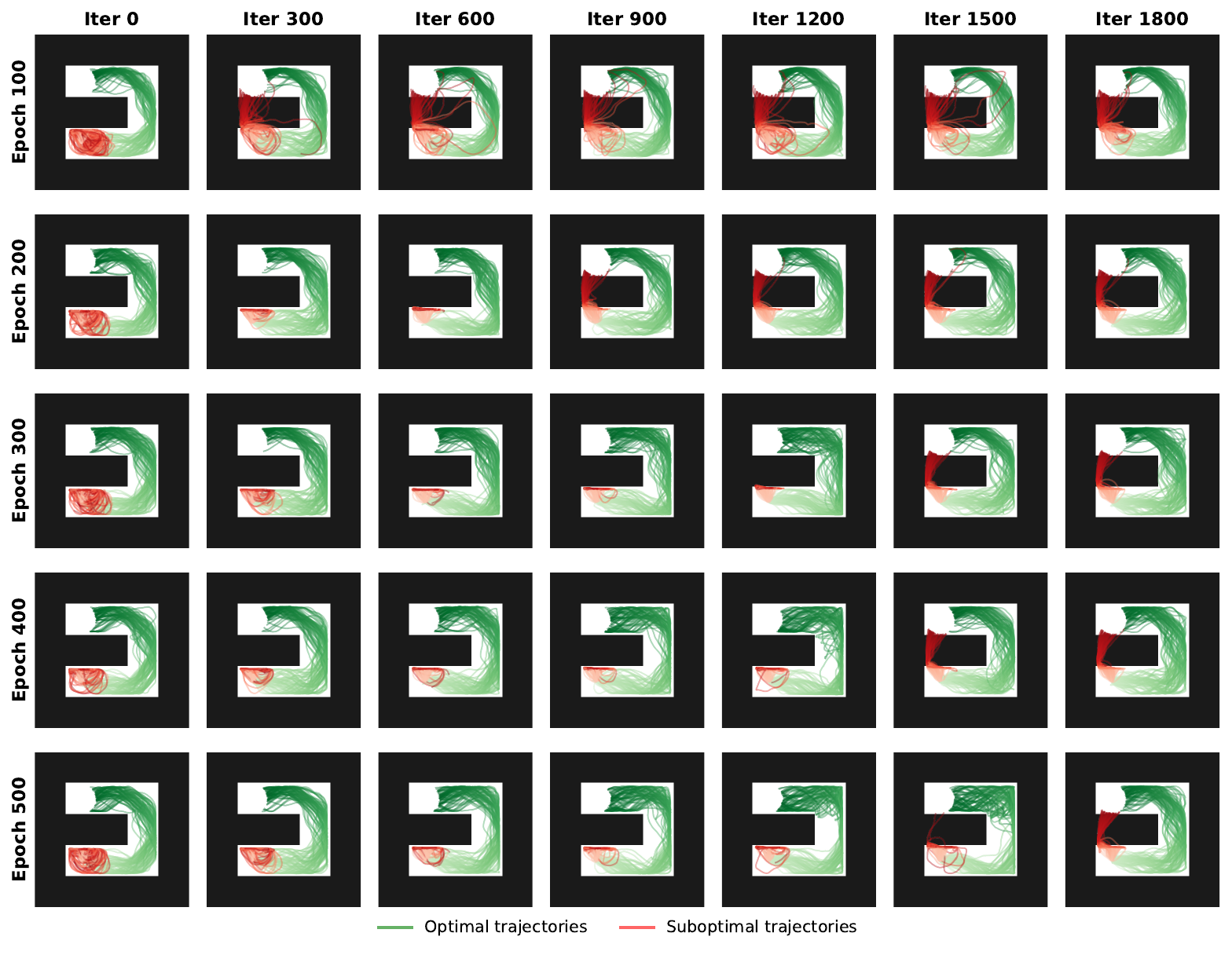}
    \caption{\textbf{Trajectory evolution during confusing instance search.} Rows show reference models of increasing quality (epochs 100-500). Columns show optimization progress (iterations 0-1800). Green: optimal policy $\pi^*$. Red: suboptimal policy $\pi_{\text{sub}}$. Better-trained models resist confusion longer, with successful suboptimal trajectories appearing much later in the optimization.}
    \label{fig:trajectory_evolution}
\end{figure}

\paragraph{How does model dynamics during the confusing instance search ?}
Figure~\ref{fig:trajectory_evolution} visualizes how the learned dynamics evolve during our optimization procedure across different model quality levels and search iterations. Each row corresponds to a reference model checkpoint (epochs 100-500), and each column shows trajectories at different points during the confusing instance search (iterations 0-1800). Green trajectories follow the optimal policy $\pi^*$, while red trajectories follow the suboptimal policy $\pi_{\text{sub}}$, both rolled out in the current confusing model candidate $M_{\tilde{\theta}}$.
At iteration 0, the confusing model is initialized to match the reference model, so $\pi_{\text{sub}}$ cannot successfully navigate through walls and remains trapped at the starting region (visible as red spiral patterns in the bottom-left corner). As optimization progresses, the dynamics are perturbed to enable wall passage for $\pi_{\text{sub}}$, indicated by red trajectories successfully reaching the goal through the middle shortcut.
The horizontal progression reveals the central finding of our work: under-trained models (epoch 100) are confused almost immediately, with successful suboptimal trajectories appearing by iteration 300. In contrast, well-trained models (epoch 500) exhibit strong resistance, requiring over 1500 iterations before $\pi_{\text{sub}}$ trajectories emerge. This visual pattern directly confirms the relationship between model quality and confusion resistance quantified by suboptimality cost.

\section{Conclusion}

We formalized the problem of finding most confusing instances for neural probabilistic world models and developed a tractable optimization framework. Our empirical investigation on VAE-based world models reveals that suboptimality cost increases with model quality, with better-trained models exhibiting stronger resistance to confusion through localized perturbations in decision-critical regions. This relationship validates $\mathbf{K}$ as a principled measure of decision-relevant uncertainty that directly connects to information-theoretic exploration requirements.
These findings open several important questions for future research. First, how can suboptimality cost drive exploration during planning? One could prioritize trajectories passing through low-$\mathbf{K}$ regions or actively seek states where computing $\mathbf{K}$ reveals high confusion potential, but translating this intuition into practical algorithms remains an open challenge. Second, how does our framework scale across different neural architectures? We focused on small VAE-based models and whether our optimization approach generalizes; architectural choices will potentially affect confusion resistance, which warrants investigation. Finally, how does the notion of confusing instances extend to partially observable environments or environments with known supports? Addressing these questions could bridge the gap between exploration theory and deep RL, potentially yielding principled, efficient exploration strategies for complex sequential decision-making systems.

\bibliographystyle{unsrtnat}
\bibliography{references}

\end{document}